\documentclass[runningheads]{llncs}
\usepackage{graphicx}
\usepackage{cite}
\usepackage{amsmath,amssymb,amsfonts}
\usepackage{algorithmic}
\usepackage{graphicx}
\usepackage{textcomp}
\usepackage{xcolor}
\usepackage{bm}
\usepackage{subfig}
\usepackage{float}
\usepackage{multirow}
\usepackage{makecell}
\usepackage{diagbox}
\usepackage{tablefootnote}
\usepackage{cleveref}
\usepackage[hyphens]{url}
\usepackage{breakurl}
\usepackage[normalem]{ulem}
\usepackage{fancyhdr}

\fancypagestyle{firststyle}
{
    \fancyhf{}
    \fancyfoot[L]{\footnotesize *This paper has been received as a research paper at CollaborateCom 2024.\\
    $^\dagger$Corresponding author}
    
}

\begin{document}
\title{BACE-RUL: A Bi-directional Adversarial Network with Covariate Encoding for Machine Remaining Useful Life Prediction*}
\titlerunning{BACE-RUL}
\author{Zekai Zhang\inst{1} \and
Dan Li$^\dagger$\inst{1}\orcidID{0000-0002-3787-1673} \and
Shunyu Wu\inst{1} \and
Junya Cai\inst{1} \and
Bo Zhang\inst{2}\orcidID{0000-0003-1129-1109} \and
See Kiong Ng\inst{3}\orcidID{0000-0001-6565-7511} \and
Zibin Zheng\inst{1}\orcidID{0000-0001-7872-7718}}
\authorrunning{Z. Zhang et al.}
\institute{School of Software Engineering, Sun Yat-Sen University, Zhuhai 528478, China 
\email{zhangzk27@mail2.sysu.edu.cn, lidan263@mail.sysu.edu.cn} 
\and School of Computer Science and Technology, China University of Mining and Technology, Xuzhou 221116, China \and
National University of Singapore, 119077, Singapore}
\maketitle              

\thispagestyle{firststyle}

\begin{abstract}
Prognostic and Health Management (PHM) are crucial ways to avoid unnecessary maintenance for Cyber-Physical Systems (CPS) and improve system reliability. Predicting the Remaining Useful Life (RUL) is one of the most challenging tasks for PHM. Existing methods require prior knowledge about the system, contrived assumptions, or temporal mining to model the life cycles of machine equipment/devices, resulting in diminished accuracy and limited applicability in real-world scenarios. This paper proposes a Bi-directional Adversarial network with Covariate Encoding for machine Remaining Useful Life (BACE-RUL) prediction, which only adopts sensor measurements from the current life cycle to predict RUL rather than relying on previous consecutive cycle recordings. The current sensor measurements of mechanical devices are encoded to a conditional space to better understand the implicit inner mechanical status. The predictor is trained as a conditional generative network with the encoded sensor measurements as its conditions. Various experiments on several real-world datasets, including the turbofan aircraft engine dataset and the dataset collected from degradation experiments of Li-Ion battery cells, show that the proposed model is a general framework and outperforms state-of-the-art methods.

\keywords{Remaining useful life prediction \and Generative models \and Prognostic and health management.}
\end{abstract}

\section{Introduction}
Prognostic and Health Management (PHM), also known as Condition-Based Maintenance (CBM), are crucial for mechanical devices to avoid unnecessary maintenance activities and improve the reliability of Cyber-Physical Systems (CPS), which is the core of ``Industry 4.0''. 
PHM comprises four stages: fault detection, fault diagnostics, fault prognostics, and decision-making \cite{medjaher2012}, among which fault prognostics is the basis of decision-making and its most challenging part is the Remaining Useful Life (RUL) Prediction \cite{zhang2018degradation}. 
The RUL of a device or system is defined as the length from the current time to the end of the machine life span \cite{si2011remaining}. RUL Prediction is the process where suitable methods are used to predict the future performance of the equipment and estimate when the equipment will break down \cite{lee2014prognostics}. Predicting RUL accurately helps implement maintenance at a proper time, preventing late or premature equipment replacement. Severe casualties and financial losses might happen if the equipment is maintained too late. However, premature maintenance leads to a waste of labor and cost when the equipment is still in good condition. 

Existing methods for RUL Prediction can be grouped into three categories, i.e., model-based, data-driven, and deep learning-based approaches. 
Model-based approaches try to model the degradation process via mathematical and physical models to predict RUL \cite{zhang2018degradation,lei2016model}. Lei et al. proposed Weighted Minimum Quantization Error (WMQE) to extract the health indicator. They initialized the model parameters using the Maximum Likelihood Estimation (MLE) algorithm and an algorithm based on Particle Filtering to predict RUL \cite{lei2016model}. Those mathematical or physical models can generate accurate results if the complex system degradation can be modeled precisely based on researchers' expertise and domain knowledge. However, accurate knowledge cannot be easily obtained for complex systems due to a lack of comprehensive understanding of such complicated structures \cite{li2018remaining}. Although considerable efforts should be made to build an explicit model, it can only be deployed for special cases or equipment with little possibility of generalization \cite{liao2014review}. 

\textcolor{black}{Data-driven methods mainly refer to those based on statistical models and traditional machine learning methods.} Benkedjouh et al. \cite{benkedjouh2013remaining} utilized the Isometric Feature Mapping Reduction (ISOMAP) and the Support Vector Regression (SVR) to learn the degradation model of bearings and predict RUL accordingly. \textcolor{black}{Kontar et al. \cite{kontar2018nonparametric} model each condition monitoring signal from the components of the equipment using Multivariate Gaussian Convolution Processes (MGCP), taking into account the heterogeneity in data. In \cite{jahani2020remaining}, B-spline basis functions are used to model and forecast degradation signals.} Si et al.  \cite{si2022nonlinear} proposed a method in which Box-Cox Transformation (BCT) was used to transform the nonlinear degradation data into nearly linear data. The Wiener Process with random drift was utilized to model the evolving process of the transformed data. 
These models are built on a basic assumption that the degradation process can be abstracted as a stochastic process and the RUL of a mechanism is a random distribution \cite{zhang2018degradation,si2011remaining,si2022nonlinear,lei2018machinery}. However, these statistical assumptions about the degradation process are not identical to the practice that a component may have multiple faulty modes \cite{si2011remaining}. 

With the development of deep neural networks, deep learning-based models have gained great interest and achieved significant results in many fields, including RUL prediction. Guo et al. \cite{guo2017recurrent} built health indicators with a Recurrent Neural Network (RNN)-based approach to prognosticate RUL by feeding the manually selected features into the RNN to build Health Indicators (HI). RUL was obtained by the downstream particle filtering algorithm using the HI \cite{qian2015remaining}. Yang et al. \cite{yang2019remaining} proposed a double-convolutional neural network architecture to predict RUL through raw sensor data without extracting health indicators beforehand. Wang et al. \cite{wang2020recurrent} developed an RNN-based RUL prediction framework by adopting a recurrent convolutional layer to model the temporal dependencies of different degradation states and employed the variational inference to quantify the uncertainty of RCNN in RUL prediction. 
However, sensor measurements collected during equipment operating cycles differ from the conventional multivariate time series data. For example, measurements from the turbofan aircraft engine dataset were collected each time the turbofan finished an operation cycle, where the time interval between two working cycles might be hours or days in real cases. Besides, real-life mechanical equipment could actively serve in factories or CPSs for decades, while most factories or CPSs could not record the life-long working conditions for each device. To be specific, for in-use devices at their late stages of life span when RUL prediction is essential, some factories can't track back to all the previous working conditions. On the other hand, for some heavy machinery, it may take months or even years to gather enough consecutive working cycles that convey enough temporal dependency. Thus, existing DL-based methods, although presenting impressive performances by adequately taking advantage of temporal dependency, are not practical enough for real-life applications.

Accordingly, a feasible RUL prediction model should meet the following criterion: it should be a general framework that could be applied to predict RUL for devices from various fields without being dependent on domain-specific knowledge and feature engineering, it should consider the distributional characteristics as much as possible without strong statistical assumptions, and it should adequately utilize the sensor measurements collected at current life cycle to capture the inner mechanical status. 
From a probabilistic perspective, the RUL of a mechanical system at a certain time could be estimated as a conditional distribution, which is conditioned on the current working status. This paper proposes a Bi-directional Adversarial network with Covariate Encoding for machine Remaining Useful Life (BACE-RUL) prediction.
An adversarially trained deep neural network is employed to estimate the distribution of the RUL conditioning on the sensor measurements without any assumption about the probability distribution. A bi-directional training mechanism is utilized, to encode the covariant sensor measurements to the conditional space to capture feature representation. The main contributions of this paper are:
\begin{itemize}
    \item BACE-RUL captures the conditional distribution of the RUL and covariant sensor measurements in a purely data-driven way, without any reliance on the assumptions of data distributions and system models;
    \item BACE-RUL predicts the RUL of any cycle in the life span of a mechanism using current sensor measurements without historical measurements from previous cycles;
    \item BACE-RUL adopts a coupled encoder-decoder architecture. The prediction process is constrained by the representations captured by encoding the ground truth data and the covariates, thus obtaining more realistic RUL representations;
    \item Experiments on several real-world datasets from different domains showed BACE-RUL's superiority over state-of-the-art methods. 
\end{itemize}

The remaining part of this work is organized as follows. 
Section~\ref{sec:model} describes the proposed BACE-RUL framework with related methodology. Experiments on three datasets to evaluate the proposed model are presented in Section~\ref{sec:Experiments}. Section~\ref{sec:conclusion} summarizes the work and proposes possible future work. 

\section{Preliminaries: System Degradation and RUL}
\label{sec:prelim}
Generally, the degradation of a system or equipment is the downgrading in performance, reliability, and remaining lifetime due to internal or external factors \cite{gorjian2010review}. As a stochastic process, the degradation process is gradual and complicated \cite{gorjian2010review,jin2019one}. Four discrete stages commonly describe the degradation process: normal degradation, transition region, accelerated degradation, and failure\cite{ramasso2014review}. The systems normally work at the beginning of their lifespan and degrade very slowly. \textcolor{black}{When health status reaches a ``change point '' in the transition region, the degradation accelerates,} and the systems degrade obviously in an approximately linear mode \cite{li2018remaining}. The systems fail when their degradation level approaches a threshold specified according to practice \cite{gorjian2010review}. 

With a slight abuse of notation, we will not distinguish the notations of RUL for systems, objects, devices, equipment, or mechanisms. We will discuss building a generalized model for accomplishing common RUL prediction tasks. The remaining useful life of an object is naturally defined as the current time till the failure. 
However, the RUL is not essential and could be very large at the beginning of their lifespan since devices are usually almost new, i.e., in the normal degradation phase. This is because the decisive factors that lead to wore-out and failure have not emerged yet, and the degradation at this stage is unclear and can be neglected to some extent \cite{severson2019}. As a result, there is no need to predict RUL very precisely at the early stage. After the transition region, the estimation of RUL becomes significant since we should be prepared for in-time maintenance at this stage. Thus, RUL values were set to be constant at the approximated normal degradation \cite{zhao2017remaining,heimes2008,li2018remaining,wu2018remaining}, and we also adopted this way to pre-process the historical data in this work (Section~\ref{subsubsec:data-pre}). 

\section{Deep Adversarial RUL Prediction}
\label{sec:model}
\subsection{System Architecture}
\label{subsec: archi}
\begin{figure}[t!]
    \centering
    \includegraphics[scale=0.32]{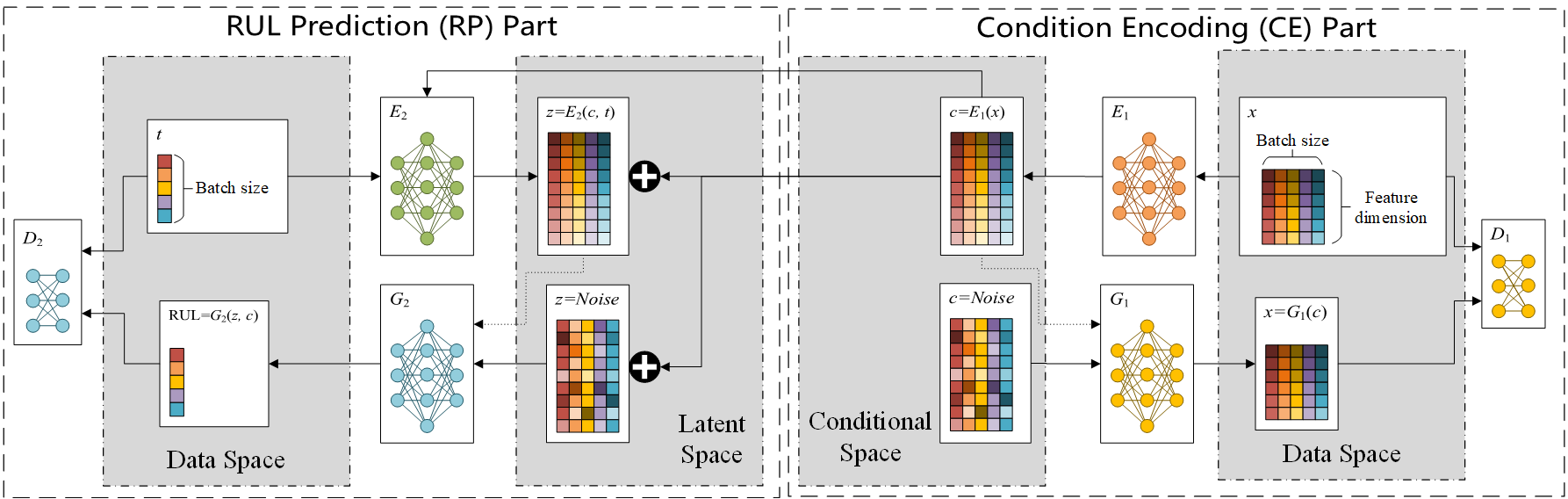}
    \caption{The proposed BACE-RUL framework.}
    \label{fig:model_structure}
\end{figure}

This work regards sensor measurements as conditions strongly correlated to the current working status. 
As shown in Fig.~\ref{fig:model_structure}, the proposed BACE-RUL model includes the Condition Encoding (CE) process in the right part and the RUL Prediction (RP) process in the left part. In the CE process, the sensor measurements $\bm{x}$ are firstly encoded by $E_1$ to a conditional space and reconstructed back to the data space by $G_1$. The conditional space's dimension is designed to be larger than the sensor measurements in the data space since we hope to extract the information better from the dimension-expanded conditional space.

In the RP process, the encoded conditions are concatenated to the random noise in the latent space and then fed to the generator $G_2$, which is implemented as an RUL predictor and projects the random noise to the data space. Besides, the current RUL $t$ is encoded to the latent space by an encoder $E_2$ on condition of $\bm{c}$ to enhance the stabilization of the generation (prediction) process.

\subsection{Methodology}
\label{subsec: method}
For the RUL prediction task, we aim to estimate $p(t|\bm{x})$, where $t$ is the current remaining life duration and $\bm{x}$ represents available sensor measurements. Here, sensor measurements $\bm{x}$ are treated as conditions of the generative model to complement the RUL prediction task. To be specific, the model's probability density function is $q_{\theta}(t|\bm{c})$, where $\bm{c}$ is mapped by a projection model $f_{\psi}(\cdot)$ which encodes the measurements $\bm{x}$ from the original space to the conditional space. 

\subsubsection{Condition Encoding.}
In the CE process, the autoencoder $E_1$ with model parameters $\bm{\psi}_{1} $ is first introduced into the architecture to map the measurements $\bm{x}$ to the conditional space,
\begin{equation}
    \label{encoder1}
    \bm{c}=E_{1}(\bm{x})
\end{equation}
where $\bm{x}\in \mathbb{R}^{m}$, $\bm{c}\in \mathbb{R}^{n} $ and $n>m$. An adversarial regenerate structure is then designed to constrain and stabilize $E_1$ by reconstructing $\bm{x}$ from $\bm{c}$,
\begin{equation}
    \label{generator1}
    \bm{x}_{recon}=G_{1}(\bm{c})
\end{equation}
where $\bm{x}_{recon} \in \mathbb{R}^{m}$ is the reconstructed sensor data. The discriminator $D_{1}$, whose parameters are denoted as $\bm{\phi}_{1}$, helps to train $G_1$ and $E_1$ adversarially by distinguishing $\bm{x}$ from $\bm{x}_{recon}$. In this stage, $E_1$, $G_1$, and $D_1$ are all designed as deep neural networks.

\subsubsection{RUL Prediction.}
In the RP process, a conditional generator $G_2$ is contrived to predict RUL by modeling $q(t|\bm{c})$. With the latent space be defined as $\bm{z} \sim p_z(\bm{z})$, the current RUL value is designed as,
\begin{equation}
    \label{generator2}
    t_{gen} = G_{2}(\bm{z}, \bm{c})
\end{equation}
where the generator $G_{2}$ is a deterministic function of $\bm{c}$ and $\bm{z}$ with model parameters $\bm{\theta}_2$, and $t_{gen}$ is the predicted RUL value which is marked as ``RUL'' in Fig.~\ref{fig:model_structure}. 

To better learn the implicit information buried under the correlations between sensor measurements and the actual RUL values, an autoencoder, $E_2$ (with model parameter  $\bm{\psi}_2$), was introduced to map the actual RUL values ($t$ as shown in Fig.~\ref{fig:model_structure}) from the data space to the latent space. 
\begin{equation}
    \label{encoder2}
    \bm{z}=E_{2} (t, \bm{c})  
\end{equation}
The discriminator for adversarially training $G_2$ and $E_2$ is denoted as $D_{2}$ to distinguish $t_{gen}$ from $t$ with model parameters be $\bm{\phi}_2$. In this process, $E_2$, $G_2$, and $D_2$ are specified as deep neural networks with corresponding model parameters. 

\subsubsection{Cost Functions.}
The CE and RP processes are trained separately in the proposed BACE-RUL framework. In the CE process, the cost functions for training $D_1$, $E_1$ and $G_1$ are formulated as follows:
\begin{equation}
\label{D1_loss}
\begin{aligned}
L_{D1} &= -\mathbb{E}_{\bm{x}\sim p}\left [ \log{D_{1}(\bm{x},E_{1}(\bm{x}) )}  \right ] \\ 
&- \mathbb{E}_{\bm{c}_{\epsilon } \sim p_{\epsilon }}\left [ \log{(1-D_{1}(G_{1}(\bm{c}_\epsilon ), \bm{c}_\epsilon))}  \right ]
\end{aligned}
\end{equation}
\begin{equation}
\label{EG1_loss}
\begin{aligned}
L_{E1,G1} &= -\mathbb{E}_{\bm{c}_{\epsilon } \sim p_{\epsilon }}\left [ \log{D_{1}(G_{1}(\bm{c}_\epsilon ), \bm{c}_\epsilon)}  \right ] \\ 
&- \mathbb{E}_{\bm{x}\sim p}\left [ \log{(1-D_{1}(\bm{x},E_{1}(\bm{x}) ))}  \right ]
\end{aligned}
\end{equation}
where $\bm{c}_{\epsilon } \sim p_{\epsilon }(\epsilon)$ is random noise sampled from $p_{\epsilon }$, $p_{\epsilon }$ is a simple probability distribution, such as an isotropic Gaussian or uniform (the latter is adopted in our study), and $p(t, \bm{x})$ represents the empirical joint distribution of data space. \eqref{D1_loss} is minimized w.r.t. parameters $\bm{\phi}_{1}$ while \eqref{EG1_loss} is minimized w.r.t. parameters $\bm{\psi}_{1}$ and $\bm{\theta}_{1}$. Besides, a reconstruction loss is defined to ensure the model keeps the essential information from $\bm{x}$ while mapping it to $\bm{c}$.
\begin{equation}
\label{recon1_loss}
    L_{recon1}=\mathbb{E}_{\bm{x}\sim p}\left \| G_{1}(E_{1}(\bm{x}))-\bm{x} \right \|_{2} 
\end{equation}
where $\left \| \cdot \right \|_{2}$ is the $l_2$ norm defined to calculate the 
distances (similarities) between the original sensor measurements $\bm{x}$ and the reconstructed measurements $G_{1}(E_{1}(\bm{x}))$ in the data space. Here, a smaller reconstruction loss ensures that $E_1$ can suitably reconstruct the information from the sensor measurement.

\textcolor{black}{Following the recent research \cite{zhao2017remaining,heimes2008,li2018remaining}, we assumed the RUL to be constant if it is larger than a threshold value at the early stage, the so-called ``RUL early constant value'', since the degradation of a system is negligible at the beginning of its lifespan, i.e. the normal degradation stages. After the normal degradation stages come the accelerated degradation stages, where the systems degrade approximately linearly \cite{li2018remaining}.} The subscripts $n$ represent data collected in the normal degradation stage while subscripts $a$ depict data collected in the accelerated degradation stages. Since the mechanism is almost new in the normal degradation stage, the actual RUL is not essential and could be very large ideally. Thus, we assume that predicted RULs in the normal degradation stage should be no less than the ground truth to capture the degradation process's multi-modal status. This means that it is not necessary to restrict the model to get very accurate predictions at an early stage. In contrast, predictions in the accelerated degradation stage are expected to be as close to the actual RULs as possible. Thus, in the RP process, we first utilized data in the accelerated degradation stage to train $D_2$, $E_2$, and $G_2$ in an adversarial manner. 
\begin{equation}
\begin{aligned}
\label{D2_loss}
 L_{D2} &= -\mathbb{E}_{(t_{a},\bm{c}_{a})\sim p_{ca}}\left [ \log{D_{2}(t_{a},E_{2}(t_{a},\bm{c}_{a}),\bm{c}_{a})}  \right ]\\
 &-\mathbb{E}_{\bm{z}_\epsilon \sim p_{\epsilon }, \bm{c}\sim p_{c}}\left [ \log{(1-D_{2}(G_{2}(\bm{z}_{\epsilon },\bm{c}),\bm{z}_{\epsilon },\bm{c}))}  \right ]    
\end{aligned}
\end{equation}
\begin{equation}
\begin{aligned}
\label{EG2_loss}
L_{E2,G2} &= -\mathbb{E}_{\bm{z}_\epsilon \sim p_{\epsilon }, \bm{c}\sim p_{c}}\left [ \log{D_{2}(G_{2}(\bm{z}_{\epsilon },\bm{c}),\bm{z}_{\epsilon },\bm{c})}  \right ]\\
 &-\mathbb{E}_{(t_{a},\bm{c}_{a})\sim p_{ca}}\left [ \log{(1-D_{2}(t_{a},E_{2}(t_{a},\bm{c}_{a}),\bm{c}_{a}))}  \right ]
\end{aligned}
\end{equation}
where $\bm{z}_{\epsilon } \sim p_{\epsilon }(\epsilon)$ is random noise sampled from $p_{\epsilon }$, which is specified as uniform distribution in our study, $p_{ca}(t_{a}, \bm{c}_{a})$ is the empirical joint distribution models the accelerated degradation stage, and $p_{c}(t, \bm{c})$ is the empirical joint distribution for all data. The expectation terms in \eqref{D2_loss} and \eqref{EG2_loss} are only computed through their corresponding samples. Here, $D_2$, $E_2$ and $G_2$ are trained by minimizing \eqref{D2_loss} w.r.t. discriminator parameters $\bm{\phi}_{2}$, and minimizing \eqref{EG2_loss} w.r.t. parameters $\bm{\theta}_{2}$ and $\bm{\psi}_{2}$ of generator and encoder, respectively. 

To ensure that the predicted RUL is as close to the ground truth as possible, we also imposed a distortion loss $L_{dist}$ for both data in the accelerated degradation stage and the normal degradation stage:
\begin{equation}
\begin{aligned}
\label{distort_loss}
L_{dist} &= \mathbb{E}_{\bm{z}_\epsilon \sim p_{\epsilon },(t_{a},\bm{c}_{a})\sim p_{ca}}\left \| t_{a}-G_{2}(\bm{z}_{\epsilon },\bm{c}_{a}) \right \|_{2} \\
&+ \mathbb{E}_{\bm{z}_\epsilon \sim p_{\epsilon },(t_{n},\bm{c}_{n})\sim p_{cn}}[\max (0, t_{n}-G_{2}(\bm{z}_{\epsilon },\bm{c}_{n})) ]^2
\end{aligned}
\end{equation}
where the first term induces predicted RUL in the accelerated degradation stage to be as close to the ground truth as possible. In contrast, the second term penalizes $G_{2}(\bm{z}_{\epsilon },\bm{c}_{n})$ for generating estimations lower than the actual RUL labels.

The reconstruction loss is also adopted to ensure that the generator $G_2$ can capture enough implicit information from the actual RUL values to make accurate predictions.
\begin{equation}
\begin{aligned}
\label{recon2_loss}
L_{recon2}=\mathbb{E}_{(t,\bm{c})\sim p_{c}}\left \| G_{2}(E_{2}(t,\bm{c}),\bm{c})-t \right \|_{2}
\end{aligned}
\end{equation}
where the actual RUL values $t$ are encoded by $E_2$ to the latent space and then reconstructed by $G_2$. By minimizing this reconstruction loss, the generator can obtain the optimal data representation of the actual RUL, which helps it with better performance and higher stability. The necessity of this reconstruction part is proved by the \textcolor{black}{research question in Section~\ref{subsec:RQ:effectiveness of coupled encoder-decoder}.}

At last, the cost functions for the proposed BACE-RUL framework are
\begin{equation}
\begin{aligned}
\label{total_loss_1}
L_{CE} = \lambda _{11} L_{D1} + \lambda _{12} L_{E1,G1} + L_{recon1}
\end{aligned}
\end{equation}
\begin{equation}
\begin{aligned}
\label{total_loss_2}
L_{RP} &= \lambda _{21} L_{D2} + \lambda _{22} L_{E2,G2} + L_{recon2} + L_{dist}
\end{aligned}
\end{equation}
where $\lambda _{11}$, $\lambda _{12}$, $\lambda _{21}$, and $\lambda _{22}$ are tuning parameters that balances each loss for the total cost functions. 

\subsubsection{Optimization.}
With one iteration, the cost function in \eqref{total_loss_1} is optimized first to train $D_1$, $E_1$ and $G_1$, and then \eqref{total_loss_2} is optimized subsequently to train $D_2$, $E_2$ and $G_2$. All the above neural networks are optimized using stochastic gradient descent on mini-batches, and the \emph{Adam} algorithm is adopted. The generator and the encoder are trained $k\enspace (k\ge 2)$ times for each cost function. Then the discriminator is trained once in an iteration since the task for the generator, and the encoder is much heavier than the discriminator. 

\begin{table}[ht]
\centering
\vspace{-20pt}
\scriptsize
\caption{Details about the datasets.}\label{tab:cmapssdetails}
\label{tab:data}
\begin{tabular}{|c|c|c|c|c|c|c|c|c|}
\Xhline{1.0pt}
 &  \multicolumn{2}{c|}{FD001} & \multicolumn{2}{c|}{FD002} & \multicolumn{2}{c|}{FD003} & \multicolumn{2}{c|}{FD004}\\
\hline
 & \makecell{training\\set} & \makecell{test\\set} & \makecell{training\\set} & \makecell{test\\set} & \makecell{training\\set} & \makecell{test\\set} & \makecell{training\\set} & \makecell{test\\set}\\
\hline
Number of engines & 100 & 100 & 260 & 259 & 100 & 100 & 248 & 249\\
\hline
Number of objects & 20631 & 13096 & 53759 & 33991 & 24720 & 16596 & 61249 & 41214\\
\hline
Conditions &  \multicolumn{2}{c|}{ONE} & \multicolumn{2}{c|}{SIX} & \multicolumn{2}{c|}{ONE} & \multicolumn{2}{c|}{SIX}\\
\hline
Fault Modes &  \multicolumn{2}{c|}{\makecell{ONE\\(HPC Degradation)}} & \multicolumn{2}{c|}{\makecell{ONE\\(HPC Degradation)}} & \multicolumn{2}{c|}{\makecell{TWO\\(HPC Degradation, \\Fan Degradation)}} & \multicolumn{2}{c|}{\makecell{TWO\\(HPC Degradation, \\Fan Degradation)}}\\
\Xhline{1.0pt}
 &  \multicolumn{4}{c|}{NASA Battery} & \multicolumn{4}{c|}{Toyota Battery}\\
\hline
 & \multicolumn{2}{c|}{training set} & \multicolumn{2}{c|}{test set} & \multicolumn{2}{c|}{training set} & \multicolumn{2}{c|}{training set}\\
\hline
Number of batteries & \multicolumn{2}{c|}{5} & \multicolumn{2}{c|}{2} & \multicolumn{2}{c|}{7} & \multicolumn{2}{c|}{3}\\
\hline
Number of objects & \multicolumn{2}{c|}{853} & \multicolumn{2}{c|}{361} & \multicolumn{2}{c|}{11402} & \multicolumn{2}{c|}{3127}\\
\hline
Conditions &  \multicolumn{4}{c|}{TWO} & \multicolumn{4}{c|}{THREE} \\
\Xhline{1.0pt}
\end{tabular}
\vspace{-20pt}
\end{table}

\section{Experiments}
\label{sec:Experiments}

In this part, we evaluate the proposed BACE-RUL by answering the following Research Questions (RQs):
\begin{itemize}
    \item \textbf{RQ1:} Is the BACE-RUL a general framework that could be applied to predict RUL for devices from various fields without being dependent on domain-specific knowledge?
    \item \textbf{RQ2:} Can BACE-RUL beat other data-driven and deep learning-based methods without relying on feature engineering?
    \item \textbf{RQ3:} Can BACE-RUL effectively predict RUL using current sensor measurements without historical measurements from previous cycles?
    \item \textbf{RQ4:} Does the conditional space and the coupled encoder-decoder architecture contribute to the effectiveness of the BACE-RUL framework?
\end{itemize}

\subsection{Datasets}

In this part, we evaluated the proposed model on three public datasets: the C-MAPSS dataset\cite{saxena2008turbofan}, NASA lithium-ion battery dataset\cite{saha2007battery}, and another battery dataset released by Toyota Research Institute\cite{severson2019}. Table~\ref{tab:cmapssdetails} shows details about the chosen datasets.

\begin{enumerate}
    \item \textbf{C-MAPSS dataset}\footnote{\url{https://data.nasa.gov/Aerospace/CMAPSS-Jet-Engine-Simulated-Data/ff5v-kuh6}}
    The C-MAPSS dataset was released by the \emph{National Aeronautics and Space Administration} (NASA) \cite{saxena2008turbofan}. The C-MAPSS dataset comprises four subsets (\texttt{FD001-FD004}), each of which was developed under different settings (from mild to harsh). 
    \item \textbf{NASA lithium-ion battery dataset}\footnote{\url{https://phm-datasets.s3.amazonaws.com/NASA/5.+Battery+Data+Set.zip}}
    This dataset is collected from the degradation experiments carried out on the battery prognostics testbed of NASA Ames Prognostics Center of Excellence (PCoE). 
    \item \textbf{Toyota Research Institute battery dataset}\footnote{\url{https://data.matr.io/1/projects/5c48dd2bc625d700019f3204}}
    First released in \cite{severson2019}, this dataset consists of measurements from 124 commercial lithium-ion phosphate (LFP)/graphite battery cells (A123 Systems, model APR18650M1A, 1.1 Ah nominal capacity). 
\end{enumerate}

\subsection{Experiment Set-up}

For aircraft engines, variables include 21 sensor measurements. For battery datasets, variables include measurements such as charge voltage, discharge voltage, and inside temperature. 

\subsubsection{Data Pre-processing.}
\label{subsubsec:data-pre}
\textcolor{black}{In the experiments, we hope not to rely much on feature engineering and expect the model to learn from the data itself, so our data pre-processing process is simple. }Since the datasets only directly marked counts of the operational cycles, we treated the remaining surviving life cycles as the RUL value at each cycle until the end-of-life time. 
Here, we let $c_i$ be the number of elapsed cycles for the $i-th$ engine or battery since it started working, then its RUL value at cycle $c$ can be calculated by
\begin{equation}
\label{get_RUL}
    {RUL}_{i,c}=C_{end} - c_i + 1\qquad i=1, \dots , n  
\end{equation}
where $C_{end}$ is the total surviving cycles of the $i-th$ engine or battery, and $n$ is the total number of engines or batteries in the dataset. 

\textcolor{black}{As mentioned in Section~\ref{sec:prelim}, the degradation of a system is negligible at the beginning of its lifespan until a ``change point" in the transition region. Following the recent research \cite{zhao2017remaining,heimes2008,li2018remaining}, we assumed the RUL to be constant if it is larger than a threshold value at the early stage. This so-called ``RUL early constant value" was set empirically in this work. }

\subsubsection{Model Parameters.}
For the C-MAPSS dataset, we used the first 70\% of the training set for training and the remaining 30\% for validation, while the test set is given separately in the dataset. For the NASA lithium-ion battery dataset, B0005, B0006, B0018, B0033, and B0036 are used for training, and B0007 and B0034 are used for testing. For the Toyota Research Institute battery dataset, we randomly chose 7 out of 10 batteries for training and the rest for testing. 

Table~\ref{tab:model_settings} shows model settings during the training processes. The generators, encoders, and discriminators are deep neural networks with different network settings. Generators and encoders are trained 10 times before discriminators are updated once in an iteration. Each deep neural network in the BACE-RUL model was optimized using stochastic gradient descent on mini-batches. The training iteration continued until $L_{recon2}$ decreased. 
All the experiments are carried out on a PC with an Intel Core i5 CPU, 16-GB RAM, and GEFORCE GTX 1650 GPU.

\begin{table}[ht]
\centering
\vspace{-20pt}
\scriptsize
\caption{Network structure and hyper-parameters.}
\label{tab:model_settings}
\begin{tabular}{|l|l|l|l|}
\hline
\diagbox{\textbf{parameters}}{\textbf{dataset}}  &  \textbf{C-MAPSS} & \textbf{NASA battery} & \textbf{Toyota battery}\\
\hline
D1 and D2 & 2 layers, [25, 25] & 2 layers, [25, 25] & 2 layers, [16, 16] \\
E1 and G1 & 3 layers, [128, 256, 128] & 3 layers, [64, 64, 64] & 3 layers, [32, 32, 32] \\
E2 and G2 & 3 layers, [50, 50, 50] & 3 layers, [32, 64, 32] & 3 layers, [32, 32, 32] \\
Conditional space dimension & 32 & 32 & 10\\
Batch size & 250 & 250 & 250\\
Learning rate & 0.001 & 0.001 & 0.001\\
Dropout rate & 0.2 & 0.2 & 0.2\\
D update$^{\mathrm{a}}$ & 1 & 1 & 1\\
G and E update$^{\mathrm{b}}$ & 10 & 10 & 10\\
RUL early constant value & 125 cycles & 125 cycles & 550 cycles\\
\hline
\multicolumn{4}{l}{$^{\mathrm{a}}$The times that discriminators update in an iteration.}\\
\multicolumn{4}{l}{$^{\mathrm{b}}$The times that generators and encoders update in an iteration.}
\end{tabular}
\vspace{-20pt}
\end{table}

\subsection{Baselines}
In this work, we compared the proposed BACE-RUL with both end-to-end approaches, regression-based models, and statistical models: 

\textcolor{black}{\emph{End-to-end models}: Long-Short-Term Memory Recurrent Neural Networks (LSTM-RNN), Deep Convolutional Neural Networks (DCNN) \cite{li2018remaining}, GAN-based DATE model \cite{chapfuwa2018}. \emph{Regression-based models}: Random Forest (RF), Bayesian Ridge Regression (BRR), Support Vector Machine (SVM). \emph{Statistical models}: Cox Proportional Hazards (Cox-PH) model and Weibull Accelerated Failure Time (Weibull AFT) model, which are used for RUL prediction in recent studies \cite{kundu2019,equeter2020}.}
Note that the LSTM-RNN and the DCNN model use time series data measurements from the current cycle and several previous consecutive cycles. \textcolor{black}{All of the above competitive methods are carefully reproduced in our experiments and the same data pre-processing is used for a fair comparison. }

\subsection{Evaluation Metrics}
The performance of our model is evaluated by the Root Mean Square Error (RMSE), scoring function, and Mean Absolute Percentage Error (MAPE). Defined in the PHM08 Challenge\cite{saxena2008damage}, the scoring function is calculated by \eqref{scoring_func}
\begin{equation}
    s=\sum_{i=1}^{n}s_i \nonumber
\end{equation}
\begin{equation}
    \label{scoring_func}
    s_i=\left\{\begin{matrix}
    e^{-\frac{d_i}{13} }-1, \quad \mathrm{for}\; d_i< 0,\\
    e^{\frac{d_i}{10} }-1, \quad \mathrm{for}\; d_i\ge 0,
    \end{matrix}\right.
\end{equation}
where $s$ denotes the score and $n$ is the total number of testing samples. $d_i=\hat{t} _i-t_i$ 
is the error between the predicted RUL value and the real one for the $i$-th testing sample. A smaller score means a better prediction. Since late predictions may cause more severe consequences than early predictions, the scoring function penalizes late predictions more than early ones.

The MAPE is defined as follows
\begin{equation}
    \label{MAPE_definition}
    MAPE(\%) = \frac{1}{n} \sum_{i=1}^{n} \left | \frac{\hat{t}_i-t_i}{t_i}  \right | \times 100\%
\end{equation}
where $t$ and $\hat{t}$ represented ground-truth values and predicted RUL values, respectively. $n$ is the number of total predictions. MAPE is more sensitive when the ground truth RUL is small. 

\textcolor{black}{When calculating the metrics, all cycles are involved, since our model is designed to predict the RUL of every life cycle.}

\subsection{RQ1: The Generality of BACE-RUL}
\label{subsec:RQ:the generality}

\begin{figure*}[ht]
    \centering
    \begin{minipage}{0.240\linewidth}
        \centering
        \subfloat[\label{fig:engine_34_FD001} E 34 of \texttt{FD001}]{\includegraphics[width=1\textwidth, height=3.0cm]{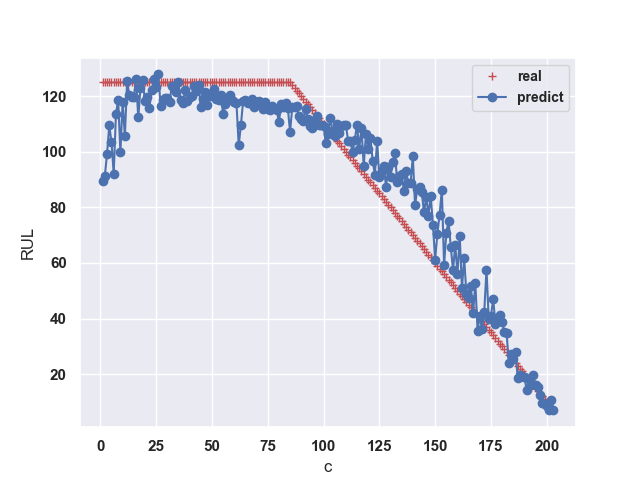}}
        \vspace{5pt}
        \subfloat[\label{fig:engine_81_FD001} E 81 of \texttt{FD001}]{\includegraphics[width=1.0\textwidth, height=3.0cm]{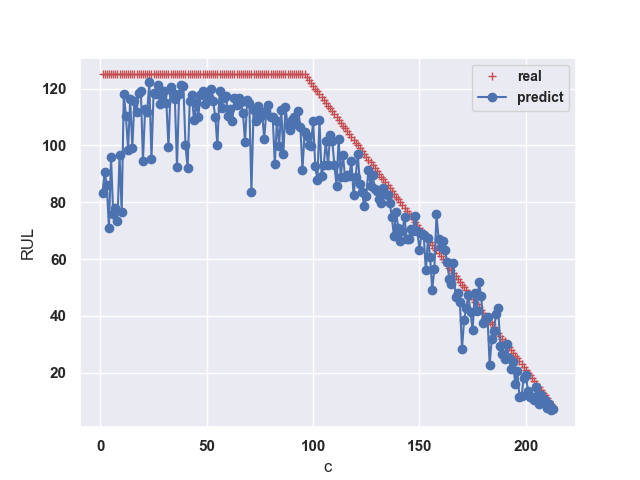}}
    \end{minipage}
    \begin{minipage}{0.245\linewidth}
        \centering
        \subfloat[\label{fig:engine_64_FD003} E 64 of \texttt{FD003}]{\includegraphics[width=1\textwidth, height=3.0cm]{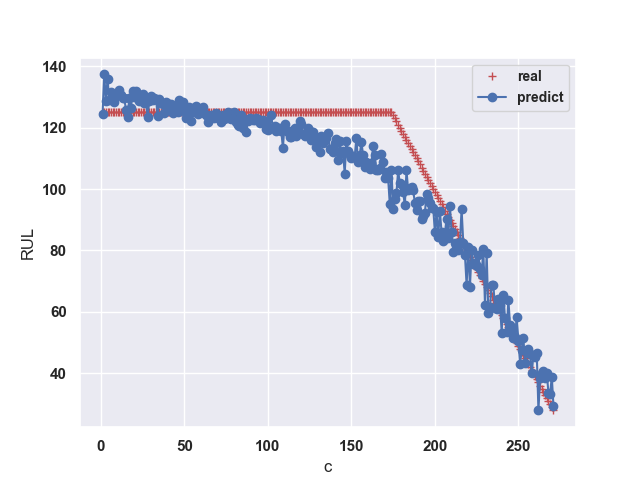}}
        \vspace{5pt}
        \subfloat[\label{fig:engine_78_FD003} E 78 of \texttt{FD003}]{\includegraphics[width=1\textwidth, height=3.0cm]{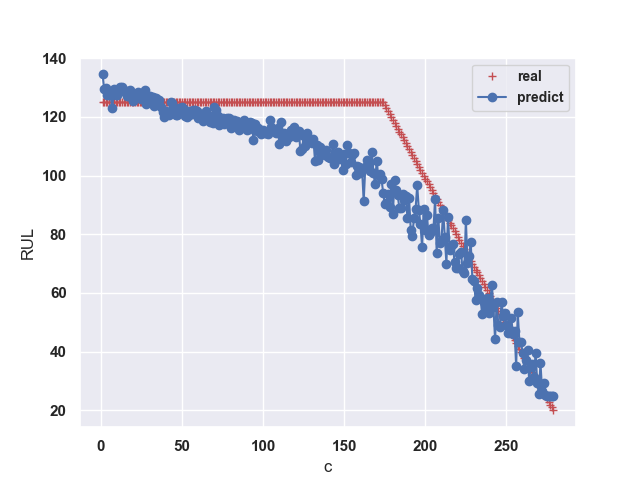}}
    \end{minipage}
    \begin{minipage}{0.245\linewidth}
        \centering
        \subfloat[\label{fig:NASA_battery_1} NASA battery 1]{\includegraphics[width=1\textwidth, height=3.0cm]{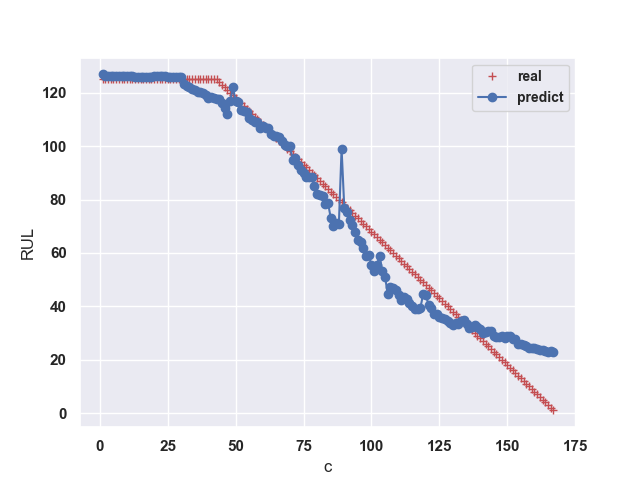}}
        \vspace{5pt}
        \hspace{5pt}
        \subfloat[\label{fig:NASA_battery_2} NASA battery 2]{\includegraphics[width=1.0\textwidth, height=3.0cm]{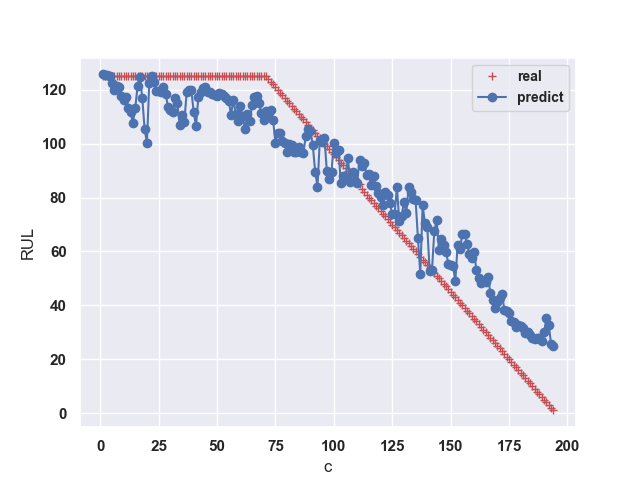}}
    \end{minipage}
    \begin{minipage}{0.245\linewidth}
        \centering
        \subfloat[\label{fig:Toyota_battery_2} Toyota battery 2]{\includegraphics[width=1\textwidth, height=3.0cm]{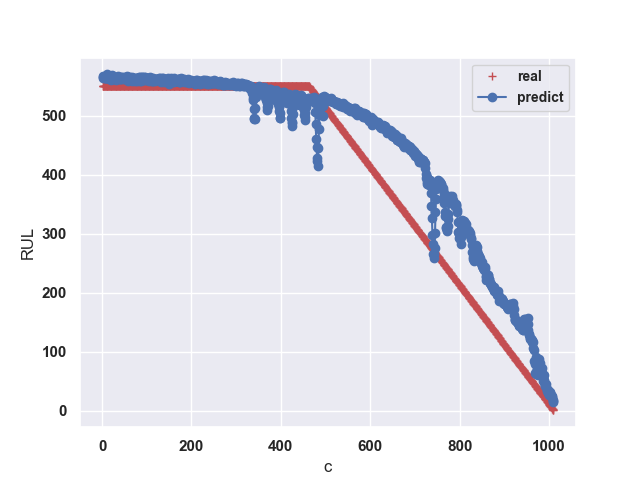}}
        \vspace{5pt}
        \hspace{5pt}
        \subfloat[\label{fig:Toyota_battery_3} Toyota battery 3]{\includegraphics[width=1.0\textwidth, height=3.0cm]{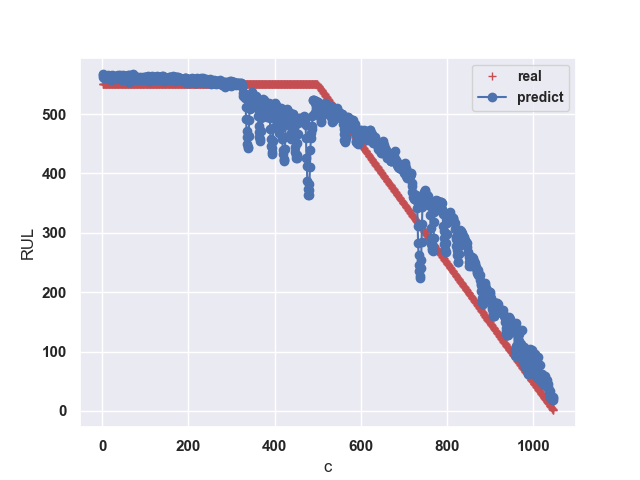}}
    \end{minipage}
    \caption{Predictions for engines and batteries during their lifespans by BACE-RUL. The horizontal axis $c$ represents the elapsed cycles from the beginning, while the vertical axis indicates the RUL at cycle $c$. ``E'' stands for ``Engine''.}
    \label{fig:predict_RUL_plot}
\end{figure*}

Fig.~\ref{fig:predict_RUL_plot} shows the predictions for the engines and batteries during their lifespans as visualization illustrations to demonstrate the efficiency of BACE-RUL. The horizontal axis $c$ represents the elapsed cycles from the engine starts working, and the vertical axis indicates the RUL of the engine at $c^{th}$ cycle. The red crosses denote ground-truth values, and the blue plots indicate predicted RUL values. \textcolor{black}{The figures show that the model performs well for both the aircraft engines and the batteries. Particularly, the BACE-RUL framework works well for the NASA Battery dataset, although there are not many training samples in this dataset. For most mechanism units, the model gives precise predictions when RUL is small.} This is crucial because humans must be alerted and prepared for maintenance when the mechanism is about to fail.

\begin{figure}[ht]
    \centering
    \subfloat[\label{fig:TSNE_FD001} T-SNE of \texttt{FD001}]{\includegraphics[width=0.25\linewidth, height=3.3cm]{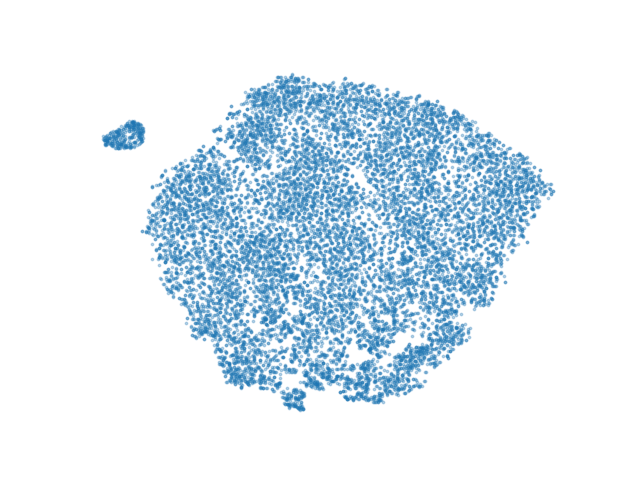}}
    \subfloat[\label{fig:TSNE_FD002} T-SNE of \texttt{FD002}]{\includegraphics[width=0.25\linewidth, height=3.3cm]{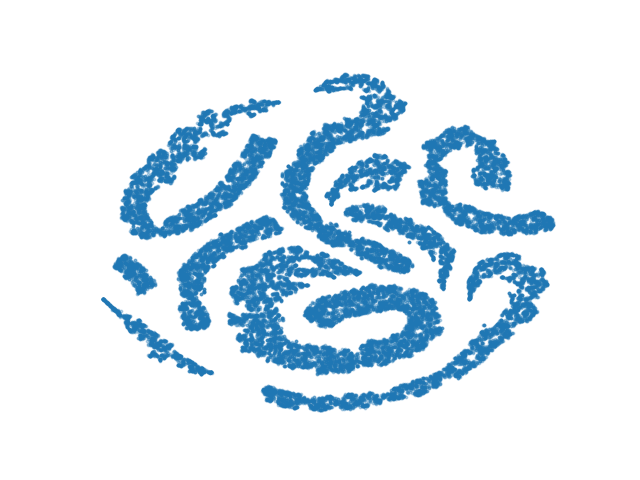}}
    \subfloat[\label{fig:TSNE_FD003} T-SNE of \texttt{FD003}]{\includegraphics[width=0.25\linewidth, height=3.3cm]{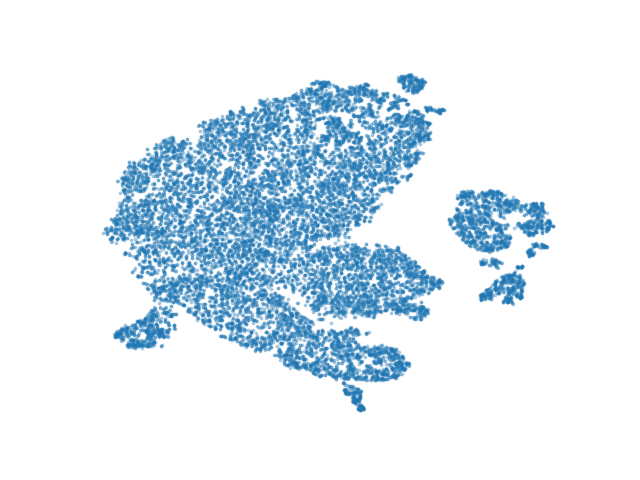}}
    \subfloat[\label{fig:TSNE_FD004} T-SNE of \texttt{FD004}]{\includegraphics[width=0.25\linewidth, height=3.3cm]{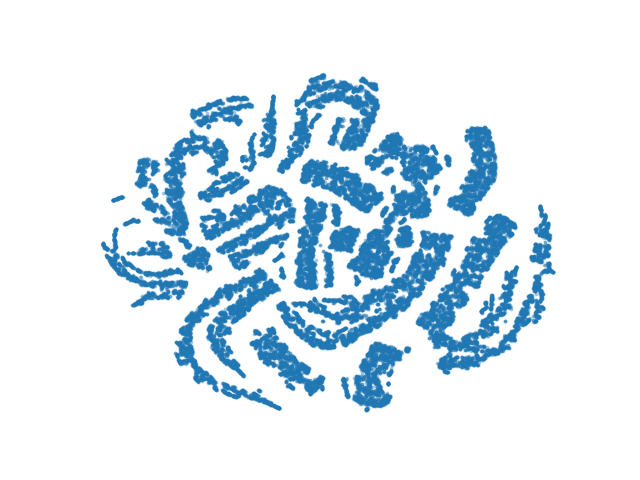}}
    \caption{T-SNE figures of four subsets of C-MAPSS. The T-SNE is a method to visualize the distribution of high-dimensional data. The more clumps in the figure indicate the more complicated the distribution is.}
    \label{fig:C-MAPSS T-SNE}
\end{figure}

\begin{figure*}[ht]
    \centering
    \begin{minipage}{0.240\linewidth}
        \centering
        \subfloat[\label{fig:FD001_81_Bayesian} E 81 of \texttt{FD001} by BRR]{\includegraphics[width=1.0\textwidth, height=3.0cm]{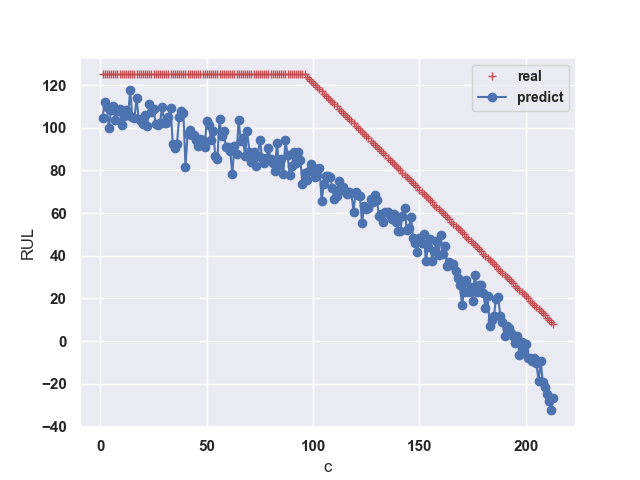}}
        \vspace{5pt}
        \subfloat[\label{fig:FD001_81_DARUL} E 81 of \texttt{FD001} by BACE-RUL]{\includegraphics[width=1.0\textwidth, height=3.0cm]{FD001_engine_81_chart.png}}
    \end{minipage}
    \begin{minipage}{0.245\linewidth}
        \centering
        \subfloat[\label{fig:FD002_76_RF} E 76 of \texttt{FD002} by RF]{\includegraphics[width=1.0\textwidth, height=3.0cm]{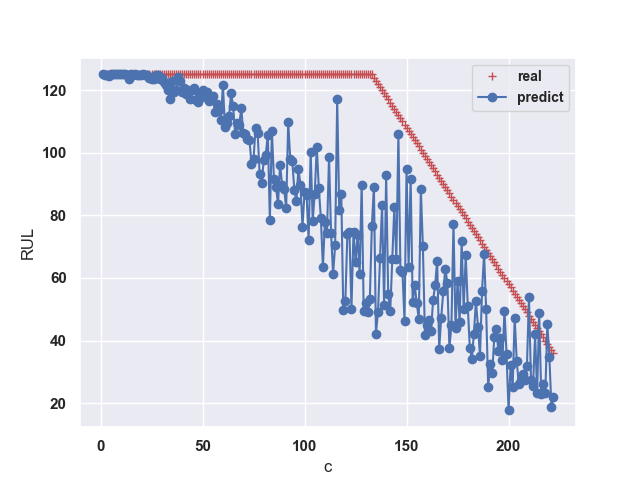}}
        \vspace{5pt}
        \subfloat[\label{fig:FD002_76_DARUL} E 76 of \texttt{FD002} by BACE-RUL]{\includegraphics[width=1.0\textwidth, height=3.0cm]{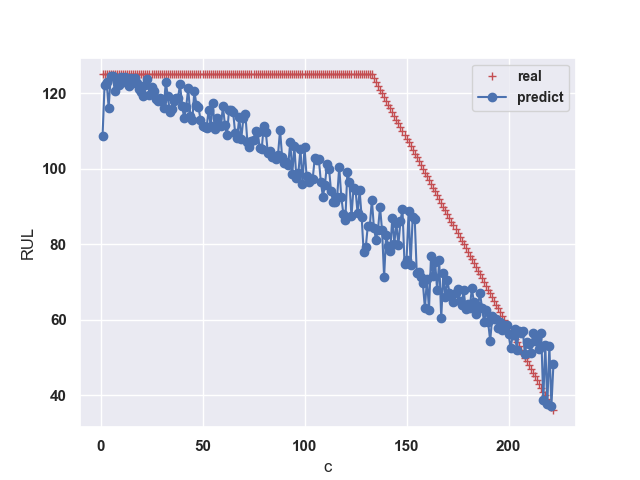}}
    \end{minipage}
    \begin{minipage}{0.245\linewidth}
        \centering
        \subfloat[\label{fig:FD004_236_DATE} E 236 of \texttt{FD004} by DATE]{\includegraphics[width=1.0\textwidth, height=3.0cm]{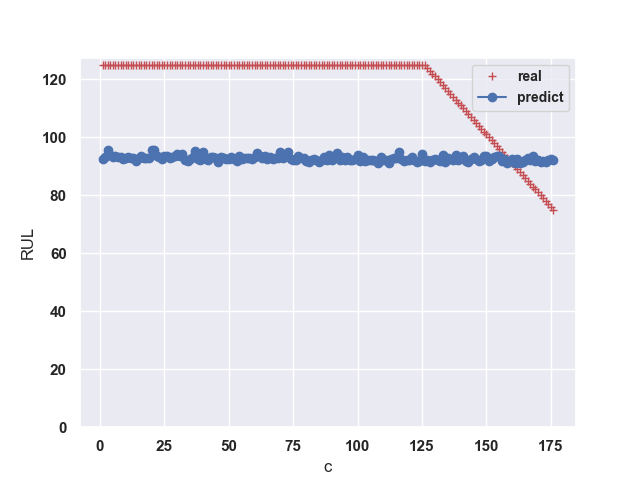}}
        \vspace{5pt}
        \subfloat[\label{fig:FD004_236_DARUL} E 236 of \texttt{FD004} by BACE-RUL]{\includegraphics[width=1.0\textwidth, height=3.0cm]{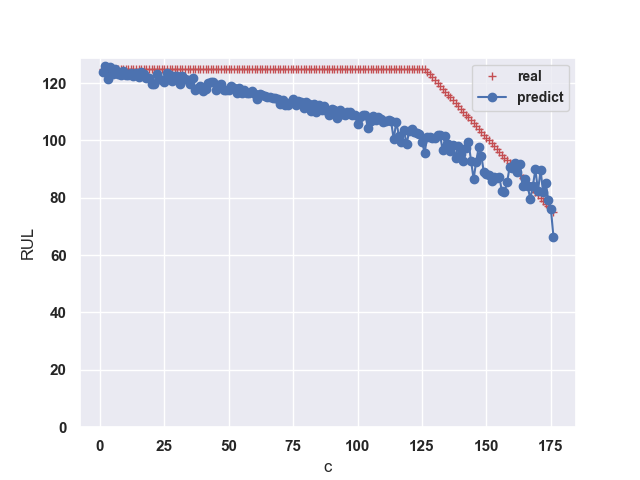}}
    \end{minipage}
    \begin{minipage}{0.245\linewidth}
        \centering
        \subfloat[\label{fig:Toyota_3_Bayesian} Toyota battery 3 by BRR]{\includegraphics[width=1.0\textwidth, height=3.0cm]{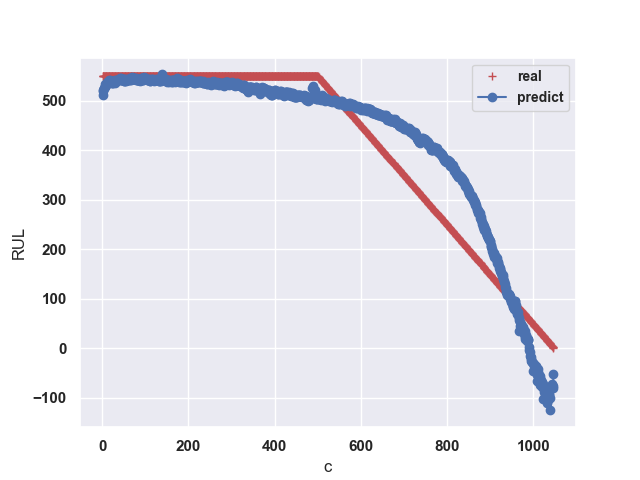}}
        \vspace{5pt}
        \subfloat[\label{fig:Toyota_3_DARUL} Toyota battery 3 by BACE-RUL]{\includegraphics[width=1.0\textwidth, height=3.0cm]{Toyota_battery_test_3_chart.png}}
    \end{minipage}
    \caption{Predictions for different engines and batteries during their lifespans using baseline methods and the proposed BACE-RUL. The horizontal axis $c$ represents the elapsed cycles from the beginning, while the vertical axis indicates the RUL at cycle $c$. ``E'' stands for ``Engine''.}
    \label{fig:predict_RUL_plot_baselines}
\end{figure*}

\subsection{RQ2: Performance of BACE-RUL Compared with Baselines}
\label{subsec:RQ:beat other methods}

We compare BACE-RUL with other baselines both visually and numerically. Only simple data pre-processing mentioned in Section~\ref{subsubsec:data-pre} is implemented without any feature engineering.

\paragraph{\textbf{C-MAPSS datasets.}}
As shown in Table \ref{tab:cmapss_comparison}, we can see that the proposed BACE-RUL model generally outperformed other baselines, especially on the \texttt{FD001} and \texttt{FD003}. For subsets \texttt{FD002} and \texttt{FD004}, \textcolor{black}{the BACE-RUL model presented the smallest or second smallest RMSE compared with other methods}, while RF and SVM achieved lower scores or MAPEs on \texttt{FD002} and \texttt{FD004}. As shown in Fig.~\ref{fig:FD002_76_RF}, with detailed screening of the RUL prediction results for engine 76 of FD002, it is seen that the RUL values of some dispersed points predicted by RF were very close to the ground truth, which resulted in lower MAPE. However, predicted RUL values by RF at many other points were far from ground truth, and the prediction curve oscillated violently since multiple fault modes exist in \texttt{FD002} and \texttt{FD004}. The T-SNE figures in Fig.~\ref{fig:C-MAPSS T-SNE} showed the complicated sub-distributions of \texttt{FD002} and \texttt{FD004} compared with \texttt{FD001} and \texttt{FD003}. This indicated that different conditions and data modes could greatly affect these traditional machine learning methods based on uniform distribution assumptions. The BRR tends to underestimate RUL (i.e., give smaller predicted RULs than ground truth), which is punished less by scoring function and gets smaller scores. Nevertheless, BRR even gets negative predicted values which are ridiculous in practice, as shown in Fig.~\ref{fig:FD001_81_Bayesian}, and this phenomenon also exists on the battery datasets  as shown in Fig.~\ref{fig:Toyota_3_Bayesian}. 

\begin{table}[htbp]
\centering
\caption{Results of different methods on C-MAPSS dataset.}
\scriptsize
\label{tab:cmapss_comparison}
\scalebox{0.93}{
\begin{tabular}{|l|l|l|l|l|l|l|l|l|l|l|}
\hline
  &  & \textbf{BACE-RUL} & LSTM & DCNN & DATE & RF & BRR & SVM & Cox-PH & Weibull AFT\\
\hline
\multirow{3}{*}{\texttt{FD001}} & RMSE & \textbf{21.81} & 31.41 & 32.76 & 29.77 & \uline{28.32} & 31.54 & 29.00 & 37.03 & 52.49\\
\multirow{3}{*}{} & Score & \textbf{72,099} & 177,115 & \uline{95,329} & 135,990 & 146,988 & 133,331 & 177,192 & 235,272 & 562,073\\
\multirow{3}{*}{} & MAPE & \textbf{15.25\%} & 37.89\% & 27.71\% & 34.58\% & \uline{20.99\%} & 26.26\% & 22.42\% & 29.11\% & 44.18\%\\
\hline
\multirow{3}{*}{\texttt{FD002}} & RMSE & \textbf{23.19} & 32.52 & 36.6 & 31.73 & 27.72 & 28.6 & \uline{23.68} & 25.31 & 44.94\\
\multirow{3}{*}{} & Score & 340,131 & 524,492 & 418,052 & 810,305 & 386,648 & \textbf{194,225} & \uline{221,693} & 304,756 & 857,530\\
\multirow{3}{*}{} & MAPE & 26.14\% & 41.55\% & 31.82\% & 40.36\% & \textbf{19.57\%} & 24.55\% & \uline{23.14\%} & 31.60\% & 36,61\%\\
\hline
\multirow{3}{*}{\texttt{FD003}} & RMSE & \textbf{29.36} & 29.79 & 34.95 & 30.96 & 34.79 & 52.55 & 36.14 & 60.32 & 73.15\\
\multirow{3}{*}{} & Score & \textbf{94,867} & 205,346 & 184,826 & 138,880 & 333,006 & 1,120,811 & 203,836 & 1,529,022 & 2,887,640\\
\multirow{3}{*}{} & MAPE & \textbf{24.11\%} & 33.46\% & 37.51\% & 32.23\% & 24.24\% & 49.99\% & 29.93\% & 51.66\% & 62,96\%\\
\hline
\multirow{3}{*}{\texttt{FD004}} & RMSE & \uline{26.41} & 28.83 & 70.26 & 32.18 & 43.29 & 38.08 & 28.76 & \textbf{24.58} & 65.06\\
\multirow{3}{*}{} & Score & 710,811 & \uline{710,122} & 8,116,539 & \textbf{709,521} & 7,941,618 & 2,089,361 & 894,535 & 1,169,113 & 5,352,168\\
\multirow{3}{*}{} & MAPE & \uline{25.79\%} & 34.34\% & 58.38\% & 35.73\% & 30.38\% & 30.67\% & \textbf{24.98\%} & 26.37\% & 52.35\%\\
\hline
\end{tabular}}
\end{table}
    
Besides, it's worth mentioning that BACE-RUL performs better than another \textcolor{black}{adversarially trained model} -- DATE by RMSE and MAPE, which is unable to learn the degradation process well from the sensor measurements and only generates values around a constant, as shown in Fig.~\ref{fig:FD004_236_DATE}. This is the advantage of the coupled encoder-decoder structure in our proposed model to help the generator get information from the training data directly and learn the complex degrade modes. DATE gets a higher prediction score on \texttt{FD004} only because there are too many cycles whose labels are equal to the ``RUL early constant value'' in the test set, as shown in Fig.~\ref{fig:FD004_236_DATE} and~\ref{fig:FD004_236_DARUL}. In this case, a lower prediction score can be produced by estimating a constant close enough to the ``RUL early constant value''. 

\paragraph{ \textbf{Battery datasets.}}
We list the results of the two battery datasets (the NASA lithium-ion battery dataset and the Toyota lithium-ion battery dataset) together in Table \ref{tab:battery_comparison}. The proposed BACE-RUL model predicted RUL effectively better than baselines by RMSE and Score on both datasets. For the MAPE, the BACE-RUL model did not beat the RF (similar results were also observed for the C-MAPSS FD002 dataset). As explained, although they presented lower mean values, results predicted by RF showed larger variances.

As shown in Table \ref{tab:data},  the size of both battery datasets is small (the training data for the NASA lithium-ion battery dataset includes 5 units with 14 features for training, and the Toyota lithium-ion battery dataset includes 7 units with 8 features for training. In comparison, the C-MAPSS dataset includes more than 100 units with 24 features for each subset.), and the dimension of variables for the NASA lithium-ion battery dataset is smaller (only 14 features) than that (24 features) of the C-MAPSS dataset. Thus, the RF method performs better on MAPE for lower prediction errors, while other deep learning-based methods tend to under-fitting.
However, experiments on battery datasets show that the proposed BACE-RUL model can generally predict RUL well with limited training data and beat other deep learning-based methods. In addition, BRR also produced negative estimated values for the battery datasets, as shown in Fig.~\ref{fig:Toyota_3_Bayesian}. 

\begin{table}[htbp]
\centering
\caption{Results of different methods on battery datasets.}
\scriptsize
\label{tab:battery_comparison}
\scalebox{0.87}{
\begin{tabular}{|l|l|l|l|l|l|l|l|l|l|l|}
\hline
  &  & \textbf{BACE-RUL} & LSTM & DCNN & DATE & RF & BRR & SVM & Cox-PH & Weibull AFT\\
\hline
\multirow{3}{*}{\texttt{NASA}} & RMSE & \textbf{11.50} & 41.21 & 41.48 & 43.12 & \uline{12.38} & 34.36 & 14.81 & 31.39 & 172.88 \\
\multirow{3}{*}{} & Score & \textbf{398} & 7,381 & 17,868 & 10,567 & \uline{401} & 11,114 & 749 & 9,958 & 5.71E+27 \\
\multirow{3}{*}{} & MAPE & \uline{63.90\%} & 172.07\% & 194.25\% & 166.87\% & \textbf{12.43\%} & 112.78\% & 88.02\% & 194.78\% & 215.82\%\\
\hline
\multirow{3}{*}{\texttt{Toyota}} & RMSE & \textbf{55.34} & 179.51 & 179.44 & 161.35 & 92.26 & 68.40 & 102.28 & 79.34 & \uline{60.40} \\
\multirow{3}{*}{} & Score & \textbf{1.69E+07} & 1.47E+18 & 3.99E+18 & 3.30E+17 & 2.67E+14 & \uline{1.64E+08} & 3.89E+12 & 1.16E+10 & 3.47E+08 \\
\multirow{3}{*}{} & MAPE & 29.21\% & 227.80\% & 234.02\% & 175.40\% & \uline{19.62\%} & 59.19\% & 88.88\% & \textbf{19.00\%} & 21.45\%\\
\hline
\end{tabular}}
\end{table}

\begin{table}[htbp]
\centering
\scriptsize
\caption{Results of the ablation studies.}
\label{tab:ablation_studies}
\begin{tabular}{|l|l|l|l|l|l|l|l|l|l|}
\hline
  & \multicolumn{3}{l|}{\textbf{Original BACE-RUL}} & \multicolumn{3}{l|}{Remove Conditional Space} & \multicolumn{3}{l|}{\textcolor{black}{Remove Reconstruction Structure}} \\
\hline
  & RMSE & Score & MAPE & RMSE & Score & MAPE & RMSE & Score & MAPE\\
\hline
\texttt{FD001} & \textbf{21.81} & \textbf{72,099} & \textbf{15.25\%} & 35.53 & 170,364 & 29.35\% & 27.94 & 195,414 & 19.65\% \\
\texttt{FD002} & \textbf{23.19} & \textbf{340,131} & 26.14\% & 28.31 & 496,102 & 33.96\% & 31.21 & 838,646 & \textbf{21.12\%} \\
\hline
\end{tabular}
\end{table}

\subsection{RQ3: The Effectiveness without Historical Measurements}
\label{subsec:RQ:effectiveness without historical}
Among the baselines, LSTM and DCNN are DL-based methods that predict RUL relying on the sensor measurements of previous consecutive life cycles (the sequence length is set to 15 for LSTM and ranges from 15 to 30 according to the original settings in \cite{li2018remaining} for DCNN). As shown in Table~\ref{tab:cmapss_comparison} and Table~\ref{tab:battery_comparison}, the proposed BACE-RUL method outperforms these two baselines. That is because it is hard for these time series-based methods to fit the training data without hand-crafted feature engineering designed for specific domains. Take datasets tested in this paper as an example, sensor measurements of aircraft engines and batteries are entirely different, they should be tailored with different feature engineering techniques. Consequently, without specific feature engineering, DCNN performs the worst on \texttt{FD002} and \texttt{FD004} due to its overfitting such complicated datasets, as the RMSEs on the training set of \texttt{FD002} and \texttt{FD004} are 11.46 and 20.60 (which are small and indicate good training performance), but 36.60 and 70.26 on the test set, respectively. On the other hand, with its ability to adequately capture inner correlations convened by covariates, the proposed BACE-RUL presents satisfactory prediction performance without complex feature engineering.

\subsection{RQ4: The Effectiveness of the Designed Model Architectures}
\label{subsec:RQ:effectiveness of coupled encoder-decoder}
For the ablation studies, some significant parts of the proposed model are removed to demonstrate its contribution to its feasibility. Two ablation strategies are designed in this work, and experiments are carried out on \texttt{FD001} and \texttt{FD002} of the C-MAPSS dataset.
\subsubsection{Remove Conditional Space.}
\label{sec:remove c space}
In the proposed model, conditional space is designed to help the model better understand the implicit system features. We remove the conditional space (as well as $G_1$, $E_1$ and $D_1$) and feed the sensor measurements $\bm{x}$ to the generator $G_2$ directly for RUL estimation in this study to give evidence of the necessity of the conditional space.
\subsubsection{Remove Encoder $E_2$.}
\label{sec:remove e2}
\textcolor{black}{In our proposed model, the advantages of the encoder-decoder architecture are adopted to improve stabilization. In this study, we try to break the reconstruction structure down by removing encoder $E_2$ during training to demonstrate the importance of the reconstruction of RUL values $t$.}

\subsubsection{Results.}
The results of the ablation studies are shown in Table~\ref{tab:ablation_studies}. We can see that removing conditional space or the reconstruction structure worsens the model's performance. Although the average MAPE by \textcolor{black}{``Remove Reconstruction Structure''} is better than the proposed BACE-RUL model shown in Table~\ref{tab:ablation_studies}, with a detailed screening of all the results, we find that it only shows good performances on limited units while behaving badly on others, which manifests instability. That is also why the accumulative prediction score is much higher than the proposed BACE-RUL model and ``Remove Conditional Space''.

\section{Conclusion}
\label{sec:conclusion}
This paper proposed a Bi-directional Adversarial network with Covariate Encoding for machine Remaining Useful Life (BACE-RUL) prediction by adopting the advantages of adversarial training and the bi-directional training mechanism. Sensor measurements were encoded into a conditional space to help the model understand the implicit, hidden information. The predictor is trained as a generative model conditioned by encoded sensor measurements. The performance of BACE-RUL is evaluated on the turbofan aircraft engine dataset and two battery datasets. Various experiments show generally lower RMSE/Score/MAPE statistics (better performance) than baseline models. 
\textcolor{black}{The proposed model is a general framework that can be used to predict RUL for different types of mechanisms without relying on historical time series data. This also makes the proposed BACE-RUL more practical in real applications. Time series change point detection can be combined to improve the accuracy of this model in the future.}

\subsubsection{Acknowledgements.} This research was founded by the Fundamental Research Funds for the Central Universities, Sun Yat-sen University (Grant No. 24QNPY147).

%
%

%
%

\begin{thebibliography}{10}
\providecommand{\url}[1]{\texttt{#1}}
\providecommand{\urlprefix}{URL }
\providecommand{\doi}[1]{https://doi.org/#1}

\bibitem{benkedjouh2013remaining}
Benkedjouh, T., Medjaher, K., Zerhouni, N., Rechak, S.: Remaining useful life estimation based on nonlinear feature reduction and support vector regression. Engineering Applications of Artificial Intelligence  \textbf{26}(7),  1751--1760 (2013)

\bibitem{chapfuwa2018}
Chapfuwa, P., Tao, C., Li, C., Page, C., Goldstein, B., Duke, L.C., Henao, R.: Adversarial time-to-event modeling. In: International Conference on Machine Learning. pp. 735--744. PMLR (2018)

\bibitem{equeter2020}
Equeter, L., Ducobu, F., Rivi{\`e}re-Lorph{\`e}vre, E., Serra, R., Dehombreux, P.: An analytic approach to the cox proportional hazards model for estimating the lifespan of cutting tools. Journal of Manufacturing and Materials Processing  \textbf{4}(1), ~27 (2020)

\bibitem{gorjian2010review}
Gorjian, N., Ma, L., Mittinty, M., Yarlagadda, P., Sun, Y.: A review on degradation models in reliability analysis. In: Engineering Asset Lifecycle Management: Proceedings of the 4th World Congress on Engineering Asset Management (WCEAM 2009), 28-30 September 2009. pp. 369--384. Springer (2010)

\bibitem{guo2017recurrent}
Guo, L., Li, N., Jia, F., Lei, Y., Lin, J.: A recurrent neural network based health indicator for remaining useful life prediction of bearings. Neurocomputing  \textbf{240},  98--109 (2017)

\bibitem{heimes2008}
Heimes, F.O.: Recurrent neural networks for remaining useful life estimation. In: 2008 international conference on prognostics and health management. pp.~1--6. IEEE (2008)

\bibitem{jahani2020remaining}
Jahani, S., Kontar, R., Zhou, S., Veeramani, D.: Remaining useful life prediction based on degradation signals using monotonic b-splines with infinite support. IISE Transactions  \textbf{52}(5),  537--554 (2020)

\bibitem{jin2019one}
Jin, B., Chen, Y., Li, D., Poolla, K., Sangiovanni-Vincentelli, A.: A one-class support vector machine calibration method for time series change point detection. In: 2019 IEEE International conference on prognostics and health management (ICPHM). pp.~1--5. IEEE (2019)

\bibitem{kontar2018nonparametric}
Kontar, R., Zhou, S., Sankavaram, C., Du, X., Zhang, Y.: Nonparametric modeling and prognosis of condition monitoring signals using multivariate gaussian convolution processes. Technometrics  \textbf{60}(4),  484--496 (2018)

\bibitem{kundu2019}
Kundu, P., Darpe, A.K., Kulkarni, M.S.: Weibull accelerated failure time regression model for remaining useful life prediction of bearing working under multiple operating conditions. Mechanical Systems and Signal Processing  \textbf{134},  106302 (2019)

\bibitem{lee2014prognostics}
Lee, J., Wu, F., Zhao, W., Ghaffari, M., Liao, L., Siegel, D.: Prognostics and health management design for rotary machinery systems—reviews, methodology and applications. Mechanical systems and signal processing  \textbf{42}(1-2),  314--334 (2014)

\bibitem{lei2016model}
Lei, Y., Li, N., Gontarz, S., Lin, J., Radkowski, S., Dybala, J.: A model-based method for remaining useful life prediction of machinery. IEEE Transactions on reliability  \textbf{65}(3),  1314--1326 (2016)

\bibitem{lei2018machinery}
Lei, Y., Li, N., Guo, L., Li, N., Yan, T., Lin, J.: Machinery health prognostics: A systematic review from data acquisition to rul prediction. Mechanical systems and signal processing  \textbf{104},  799--834 (2018)

\bibitem{li2018remaining}
Li, X., Ding, Q., Sun, J.Q.: Remaining useful life estimation in prognostics using deep convolution neural networks. Reliability Engineering \& System Safety  \textbf{172},  1--11 (2018)

\bibitem{liao2014review}
Liao, L., K{\"o}ttig, F.: Review of hybrid prognostics approaches for remaining useful life prediction of engineered systems, and an application to battery life prediction. IEEE Transactions on Reliability  \textbf{63}(1),  191--207 (2014)

\bibitem{medjaher2012}
Medjaher, K., Tobon-Mejia, D.A., Zerhouni, N.: Remaining useful life estimation of critical components with application to bearings. IEEE Transactions on Reliability  \textbf{61}(2),  292--302 (2012)

\bibitem{qian2015remaining}
Qian, Y., Yan, R.: Remaining useful life prediction of rolling bearings using an enhanced particle filter. IEEE Transactions on Instrumentation and Measurement  \textbf{64}(10),  2696--2707 (2015)

\bibitem{ramasso2014review}
Ramasso, E., Saxena, A.: Review and analysis of algorithmic approaches developed for prognostics on cmapss dataset. In: Annual Conference of the Prognostics and Health Management Society 2014. (2014)

\bibitem{saha2007battery}
Saha, B., Goebel, K.: Battery data set. NASA AMES prognostics data repository  (2007)

\bibitem{saxena2008turbofan}
Saxena, A., Goebel, K.: Turbofan engine degradation simulation data set. NASA Ames Prognostics Data Repository  \textbf{18} (2008)

\bibitem{saxena2008damage}
Saxena, A., Goebel, K., Simon, D., Eklund, N.: Damage propagation modeling for aircraft engine run-to-failure simulation. In: 2008 international conference on prognostics and health management. pp.~1--9. IEEE (2008)

\bibitem{severson2019}
Severson, K.A., Attia, P.M., Jin, N., Perkins, N., Jiang, B., Yang, Z., Chen, M.H., Aykol, M., Herring, P.K., Fraggedakis, D., et~al.: Data-driven prediction of battery cycle life before capacity degradation. Nature Energy  \textbf{4}(5),  383--391 (2019)

\bibitem{si2022nonlinear}
Si, X.S., Li, T., Zhang, J., Lei, Y.: Nonlinear degradation modeling and prognostics: A box-cox transformation perspective. Reliability Engineering \& System Safety  \textbf{217},  108120 (2022)

\bibitem{si2011remaining}
Si, X.S., Wang, W., Hu, C.H., Zhou, D.H.: Remaining useful life estimation--a review on the statistical data driven approaches. European journal of operational research  \textbf{213}(1),  1--14 (2011)

\bibitem{wang2020recurrent}
Wang, B., Lei, Y., Yan, T., Li, N., Guo, L.: Recurrent convolutional neural network: A new framework for remaining useful life prediction of machinery. Neurocomputing  \textbf{379},  117--129 (2020)

\bibitem{wu2018remaining}
Wu, Y., Yuan, M., Dong, S., Lin, L., Liu, Y.: Remaining useful life estimation of engineered systems using vanilla lstm neural networks. Neurocomputing  \textbf{275},  167--179 (2018)

\bibitem{yang2019remaining}
Yang, B., Liu, R., Zio, E.: Remaining useful life prediction based on a double-convolutional neural network architecture. IEEE Transactions on Industrial Electronics  \textbf{66}(12),  9521--9530 (2019)

\bibitem{zhang2018degradation}
Zhang, Z., Si, X., Hu, C., Lei, Y.: Degradation data analysis and remaining useful life estimation: A review on wiener-process-based methods. European Journal of Operational Research  \textbf{271}(3),  775--796 (2018)

\bibitem{zhao2017remaining}
Zhao, Z., Liang, B., Wang, X., Lu, W.: Remaining useful life prediction of aircraft engine based on degradation pattern learning. Reliability Engineering \& System Safety  \textbf{164},  74--83 (2017)

\end{thebibliography}
\end{document}